\documentclass[a4paper,11pt]{ijamas}

\usepackage{amsmath}
\usepackage{tikz}
\usepackage{pgfplots}
\usepackage{makecell}
\usepackage{hyperref}

\begin{document}


\title{Indonesian Text-to-Image Synthesis with Sentence-BERT and FastGAN}

\author{\textbf{Made Raharja Surya Mahadi$^1$ and Nugraha Priya Utama$^1$}}

\date{$^1$School of Engineering and Informatics Institute Technology Bandung\\
23520022@std.stei.itb.ac.id\\
utama@informatika.org\\[0.3cm]}

\maketitle


\begin{abstract}
\noindent \emph{Currently, text-to-image synthesis uses text encoder and image generator architecture. Research on this topic is challenging. This is because of the domain gap between natural language and vision. Nowadays, most research on this topic only focuses on producing a photo-realistic image, but the other domain, in this case, is the language, which is less concentrated. A lot of the current research uses English as the input text. Besides, there are many languages around the world. Bahasa Indonesia, as the official language of Indonesia, is quite popular. This language has been taught in Philipines, Australia, and Japan. Translating or recreating a new dataset into another language with good quality will cost a lot. Research on this domain is necessary because we need to examine how the image generator performs in other languages besides generating photo-realistic images. To achieve this, we translate the CUB dataset into Bahasa using google translate and manually by humans. We use Sentence BERT as the text encoder and FastGAN as the image generator. FastGAN uses lots of skip excitation modules and auto-encoder to generate an image with resolution $512 \times 512 \times 3$, which is twice as bigger as the current state-of-the-art model \cite{Zhang2019}. We also get $4.76 \pm 0.43$ and $46.401$ on Inception Score and Fréchet inception distance, respectively, and comparable with the current English text-to-image generation models. The mean opinion score also gives as $3.22$ out of 5, which means the generated image is acceptable by humans.} Link to source code:  \href{https://github.com/share424/Indonesian-Text-to-Image-synthesis-with-Sentence-BERT-and-FastGAN}{https://github.com/share424/Indonesian-Text-to-Image-synthesis-with-Sentence-BERT-and-FastGAN}

\medskip


\medskip

\end{abstract}


\section{Introduction}

Text-to-Image generation is challenging task because there is a domain gap between natural language and vision. Nowadays, researchers use text encoder and image generator architecture to produce photo-realistic images \cite{Reed2016,Zhang2019,Qiao2019,Tsue2020,Hu2021,Xu2018}. They use CNN-LSTM as the text encoder and Generative Adversarial Networks proposed by \citeasnoun{goodfellow14} as the image generator.

\bigskip

Lots of research in this area focus on how to generate a photo-realistic image, but on the other hand, research on the language area is rarely done. There is a thousand language across the world. One of the popular languages in South-East Asia is Bahasa Indonesia. Bahasa Indonesia is the official language of Indonesia and has been taught in the Philippines, Australia, and Japan. This language is also used in other countries such as Canada, Vietnamese, and Ukraine.

\bigskip

The main problem why research in this area is rarely done is because translating or creating a new dataset with good quality for those languages needs much cost. To examine how well text-to-image generation models perform in other languages, we translate the dataset using Google Translate. To increase the translation result, we also manually translate some of them.

\bigskip

To maintain the resolution of the generated image, we use FastGAN \cite{Liuf2021} as the image generator, which is can generate high-resolution image and comparable with StyleGAN2 \cite{Karras2020}. FastGAN proposed a skip excitation module, which is a skip connection layer. This architecture can generate a high-resolution image in just minimal iteration. The discriminator used auto-encoder-like architecture to distinguish real and fake images. To get better results, we also use Sentence-BERT \cite{Reimers2020} as the text encoder. Sentence-BERT can produce text embedding with lower cosine similarity on similar semantic sentence and vice versa.


\section{Related Works}

Nowadays, text-to-image generation uses a text encoder and image generator architecture. \citeasnoun{Mansimov2016} first proposed a method that uses deep learning, where LSTM as the text encoder and DRAW as the image generator. This method will generate an image by patches based on the attended features on the text. Next, \citeasnoun{Reed2016} uses Generative Adversarial Networks \cite{goodfellow14} as the image generator, and CNN-LSTM as the text encoder. The text encoder generates text embedding and concatenates it with random vector $Z\sim\mathcal{N}(0, 1)$. This latent vector is also used as input in the discriminator, making the discriminator distinguish wrong images.

\bigskip

In order to generate high-resolution image, \citeasnoun{Zhang2019} uses a stack of GAN to generate an image from low features to higher features. They proposed Conditioning Augmentation networks to generate smooth latent vectors and small random perturbations to increase the generated images variations. They also proposed a JCU discriminator to perform the unconditional and conditional tasks. The unconditional GAN discriminator can only distinguish between real and fake images, and the conditional GAN discriminator can distinguish real, fakes, and wrong images. They use three stacks of GAN: the first GAN generates $64 \times 64$ images, the second GAN generates $128 \times 128$ images, and the last GAN generates $256 \times 256$. This method becomes the first and state-of-the-art photo-realistic text to image generation.

\bigskip
\citeasnoun{Xu2018} proposed Attention GAN to improve the generated image using word-level attention. They proposed a new Deep Attentional Multimodal Similarity Model (DAMSM) to perform the attention. While \citeasnoun{Qiao2019} use different approaches by improving the learning text-to-image generation idea using their proposed new module, Semantic Text Module (STEM), Global-Local collaborative Attentive Module in Cascaded Image Generators (GLAM), and Semantic Text Regeneration and Alignment Module (STREAM). STEM perform text embedding using global sentence features and word-level features. GLAM is a multi-stage Generator to generate a realistic image. STREAM is an image captioning module that semantically aligns with the given text descriptions.

\bigskip

In the same year, \citeasnoun{Qiao22019} also proposed a new method to perform text-to-image generation, which is inspired by how humans draw a picture from a given description. To mimic this process, they propose LeicaGAN, which consists of three-phase, firstly is multiple priors learning via Textual-Visual Co-Embedding (TVE), next is Imagination via Multiple Priors Aggregations (MPA), and the last one is Creation via Cascaded Attentive Generator (CAG). TVE is a module that generates text embedding with the same common semantic space as an image. MPA is used to imagine what image will be generated. The text embedding and text mask from the TVE module are used to convey visual information about the semantics, textures, colors, shapes, and layouts. CAG module is the actual generator module to produce the image.

\bigskip

To improve the attention results from \citeasnoun{Xu2018}, \citeasnoun{Tsue2020} use BERT \cite{Devlin2019} as the text encoder and CycleGAN \cite{Zhu2017} as the Generator. They proposed a new cyclic design to learn and map the generated image back to text descriptions. This method increases the inception score significantly.

\bigskip

Currently, to generate high-resolution images, researchers use a multi-stage generator. This makes the model size is significantly huge. \citeasnoun{Liu2021} proposed new lightweight GAN architecture that can be trained on a small amount of image and minimum computing cost to generate a high-resolution image on $1024 \times 1024$. They use skip-layer channel-wise excitation module and an auto-encoder-like discriminator to generate a high-resolution image that can be compared to the state-of-the-art StyleGAN2 \cite{Karras2020}.

\bigskip

Many research focuses on generating high-resolution images and semantically correct with the given text description. But research on the text encoder and other languages is necessary. CNN-LSTM has become popular text embedding since \citeasnoun{Reed2016} use that method. Then \citeasnoun{Xu2018} improve the text encoder with word-level attention to improve the generated image details. And then \citeasnoun{Tsue2020} use BERT as text encoder and word embedding. One of the BERT variants that is suitable for sentence embedding is Sentence BERT \cite{Reimers2020}. This variant uses a siamese architecture network to produce text embedding that can then be compared with cosine similarity to find sentence with similar meanings.


\section{Dataset}
In this research, we only use CUB dataset \cite{WahCUB_200_2011} and translated into Bahasa Indonesia using google translate and manually by humans. This dataset contains 200 birds species and almost 12k images. Each image has $10$ captions in English. We split the dataset into $8.855$ as training data and $2.933$ as validation data. This makes the total captions for training data is $88.550$ and $29.330$ for the validation data. Because we use google translate as the translation tool, we cannot expect the translation results grammarly correct. So we were trying to fix this as much as possible manually.

\section{Method}
\subsection{Text Encoder}
\label{section_text_encoder}
To generate an image from a given text description, we need to extract the feature from the given text and use that feature as input to the image generator. The easiest way to achieve this is to use text embedding. Current state-of-the-art language modeling is BERT \cite{Devlin2019}, and their variant for generating text embedding is Sentence BERT \cite{Reimers2020}. This architecture uses a siamese network to perform the training process.

\begin{figure*}
	\centering
		\includegraphics[scale=0.6]{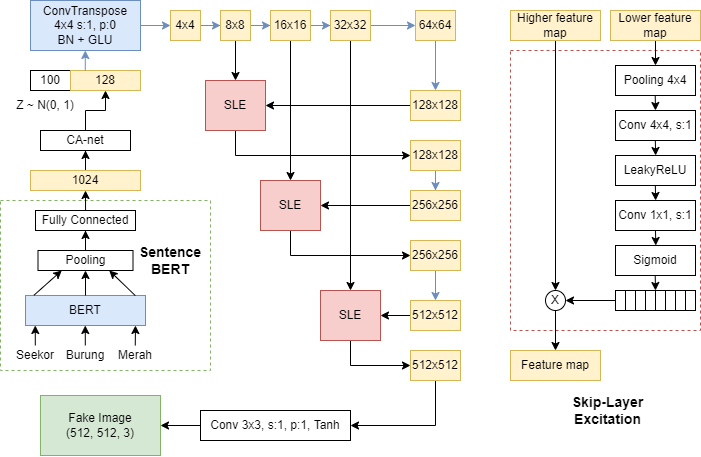}
	\caption{Our Generator Architecture. The blue box and arrow represent the same up-sampling structure, the Yellow box represents the features map on the spatial size, and the red box represents the skip-layer excitation module.}
	\label{fig_generator}
\end{figure*}

\bigskip

Usually, BERT produces a sequence of word features. The easiest way to generate a single fixed-vector as the text embedding is to feed the output into the pooling layer. As seen in the figure \ref{fig_generator}, the output from the BERT is forwarded into the pooling layer to generate a fixed vector. We can feed the feature into fully connected layers to generate different vector sizes.

\bigskip

To train this module, first, we need to pre-train the BERT. Currently, there are state-of-the-art pre-trained BERT models in Bahasa Indonesia trained by \cite{cahya} using 522MB of Indonesian Wikipedia. This model is uncased and trained under Masked Language Modeling (MLM) objective. The generated word features vector shape is $786$. We increase the output size to $1024$ using a fully connected layer.

\bigskip

Next, fine-tune the model using siamese network architecture. The loss function plays an important role here because it can determine how well the model performs on a specific task. We use cosine similarity loss with $[0.8, 1)$ on positive pair and $[0.4, 0.6]$ on negative pair. Sentence with same image class is positive pair and vice versa. Because every sentence on the dataset talks about birds, we did not use the non-zero label on the negative pair.

\subsection{Image Generator}
To perform image generation, we need a model that can generate images from given features vector from the text encoder. To achieve this, we can use Generative Adversarial Networks (GAN) \cite{goodfellow14}. GAN can generate high-resolution images from a single latent vector. To build GAN architecture, we need a generator and discriminator. Generator and discriminator need input from the text encoder to distinguish real, fake, and wrong images.

\bigskip

In order to generate high-resolution images, we use FastGAN \cite{Liuf2021} as the image generator. We can train this model with minimal effort to generate high-resolution images. Data augmentation plays an important role in this architecture.

\subsubsection{Generator}
\citeasnoun{Liuf2021} proposed a novel skip-layer excitation module with reformulate the skip-connection idea from ResNet \cite{He2016}. They use channel-wise multiplication between the activations to reduce the computational cost. Since the channel-wise multiplication does not require equal spatial dimension, we can perform a skip-connection layer between longer range resolutions, making a long shortcut to the gradient flow.

\bigskip

Skip excitation module can be defined as:
\begin{equation}
    y = F(x_{low}, \{W_{i}\}) \cdot x_{high}
\end{equation}
Where x is the input feature, and y is the output feature-maps, the function $F$ performs the operation on the lower feature of $x$, and $W_{i}$ is the weights to be learned.


\bigskip

Figure \ref{fig_generator} illustrate our Generator architecture with the output image is $512 \times 512 \times 3$. In order to generate diversity image from a single text description, We use the conditioning variable $\hat{c}$ from CA-net \cite{Zhang2019} and concatenate it with the random vector $z \thicksim \mathcal{N}(0, 1)$. To obtain $\hat{c}$, we can use equation \ref{formula_ca_net}

\begin{equation}
\label{formula_ca_net}
    \hat{c} = \mu + \sigma \odot \omega
\end{equation}

Feed the text embedding $\varphi_{t}$ into a fully connected layer to obtain $\mu$ and $\sigma$ (first-half element is $\mu$ and the rest is $\sigma$). The $\odot$ symbol is element-wise multiplication. The output dimension of the $\hat{c}$ is $128$, and the $z$ dimension is $100$, which makes our latent vector dimension is $228$. We find that normal distribution performs better than the uniform distribution for generating $z$ vector.

\subsubsection{Discriminator}
In order to perform strong regularization of the discriminator, we can treat the discriminator as auto-encoder-like architecture. There are differences with typical auto-encoder architecture, where the discriminator only decodes the image for real images on small resolution. The discriminator also performs a random crop with $\frac{1}{8}$ of its height and width on both input real images and the generated image from the decoder. The decoder only consists of $4\times$ nearest up-sampling layer, $3\times3$ convolution layer, Batch Normalization, and GLU activation function. This technique makes the discriminator learn how to reproduce the input image. 

\bigskip

\begin{figure*}
	\centering
		\includegraphics[scale=0.5]{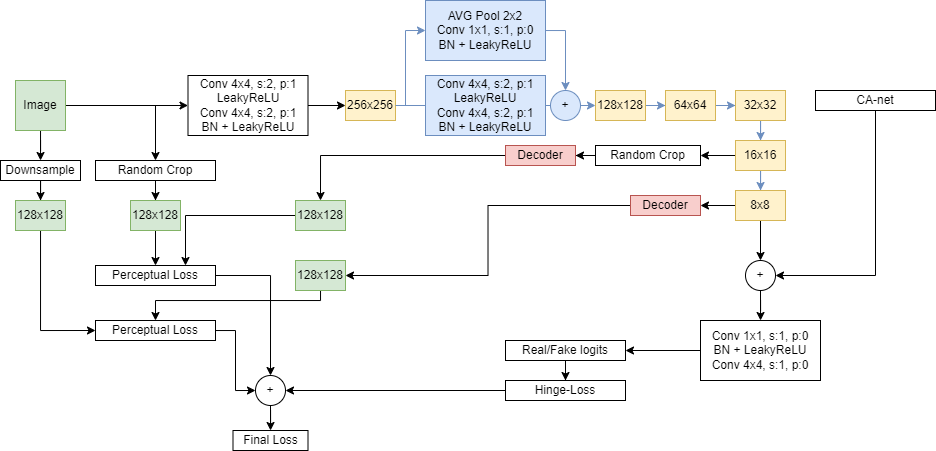}
	\caption{Our Discriminator Architecture. Blue box and arrow represent the same down-sampling structure, Yellow box represent the features map on the spatial size, and red box represent the decoder module}
	\label{fig_discriminator}
\end{figure*}

In order to perform conditional GAN, we use the $\mu(\varphi_{t})$ from the CA-net and concatenate it with the extracted feature from the discriminator, where $\varphi_{t}$ is the text embedding. To calculate the total loss of this architecture, we divide it into two conditions. The first is when the input is the real images. We use perceptual loss \cite{Zhang2018} to the auto-encoder-like architecture and hinge adversarial loss \cite{lim2017geometric} to the discriminator output and sum the total. And the last one is when the input is the fake or wrong image, we only use the hinge adversarial loss from the generator output.

\bigskip

We can define the loss function for the generator and discriminator as:
\begin{equation}
\label{eq_loss}
\begin{split}
    \mathcal{L}_{percept} &= \mathbb{E}_{f\thicksim D_{encode}(x), x\thicksim I_{real}}[\|\mathcal{G}(f) - \mathcal{T}(x) \|] \\
    \mathcal{L}_{D} &= -\mathbb{E}_{x\thicksim I_{real}}[min(0, -1 + D(x, \mu))] \\
    &\quad - \mathbb{E}_{x'\thicksim I_{wrong}}[min(0, -1 -D(x', \mu))] \\
    &\quad - \mathbb{E}_{\hat{x}\thicksim G(z, \hat{c})}[min(0, -1 -D(\hat{x}, \mu))] \\
    &\quad + \mathcal{L}_{percept} \\
    \mathcal{L}_{G} &= -\mathbb{E}_{z\thicksim\mathcal{N}}[D(G(z, \hat{c}), \mu)]
\end{split}
\end{equation}

where $\mathcal{L}_{percept}$ is the perceptual loss from \cite{Zhang2018}, $\mu$ is the variational text embedding from equation \ref{formula_ca_net}.

\section{Experiment}
\subsection{Training Details}
In this section will explain the training details for both text encoder and image generator in this research.

\subsubsection{Text Encoder}
In order to train Sentence BERT, we need pre-trained BERT models. In this research, we use \cite{cahya} pre-trained BERT on Bahasa Indonesia language. We need appropriate labels for both positive and negative pairs to get better results. In the section \ref{section_text_encoder}, we explain that we use $[0.8, 1)$ as the positive pair label and $[0.4, 0.6]$ as the negative pair labels. To train Sentence BERT, we pass both sentences in the pair to the same networks, compare them with cosine similarity, and perform mean-squared error to calculate the loss value. We train our models for 10 epochs and save the best model only.

\subsubsection{Image Generator}
We perform both unconditional and conditional tasks to compare the results. We train our models within $50.000$ iterations and batch size 10. First, we encode all text descriptions into their embedding vector $\varphi_{t}$.

\bigskip

\textbf{Unconditional}. In this part, we only feed the generator with random vector $z \thicksim \mathcal{N}(0, 1)$ and conditioning text embedding $\hat{c}$. Because every image has 10 captions, we select one randomly and feed it into CA-net. The real and fake images are then augmented with random color brightness, saturation, contrast, and translation. Next, we calculate the loss value to update the discriminator. For the real images, we perform both perceptual and hinge loss from equation \ref{eq_loss} (without the wrong image part).

\bigskip

\textbf{Conditional}. We perform the same things with the generator on the unconditional task for this task. We feed real, wrong, and fake images with the conditioning augmentation vector $\mu$ from the CA-net in the discriminator. We calculate the loss using the equation \ref{eq_loss}.

\subsection{Evaluation}
To evaluate generative models is quite challenging because calculating the suitability of the resulting image with the given text is challenging to do. So we perform both quantitative and qualitative evaluation metrics.

\bigskip

\textbf{Inception Score}. This method calculates the distribution of the generated images. It makes us can calculate how the objectness of the generated image. Inception score is often used as an evaluation metric on generative models.

\bigskip

\textbf{Fréchet inception distance}. On the other hand, Fréchet inception distance \cite{Heusel2017} has a better approach. This method calculates the distance between training and fake images data distribution. However, this method still cannot evaluate how appropriate the generated image is with the given text description.

\bigskip

\textbf{Mean Opinion Score}. In order to perform qualitative evaluation, we evaluate our models by conducting a survey. The respondent will be given 10 pair of fake images and their text description. Every fake image consists of 4 generated images. They will select one of them and then do scoring between 0 and 5. While 0 means the generated image is not appropriate with the given text description and the image is not clear, and 5 means the generated image is appropriate with the given text description and the generated image is clear.

\section{Results}
We train both text encoder and image generator separately and then combine them on evaluation. To investigate our models, we also compare them with current state-of-the-art text-to-image synthesis models to examine how the text encoder performs in a different language, especially Bahasa Indonesia.

\bigskip

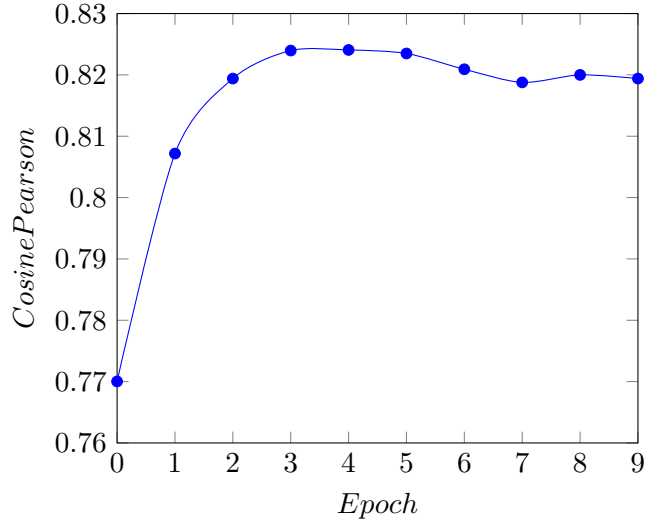
\begin{figure}
	\centering
	\begin{tikzpicture}
    	\begin{axis}[
        xlabel=$Epoch$,
        ylabel=$Cosine Pearson$,
        xmin=0, xmax=9,
        ymin=0.76, ymax=0.83,
        xtick={0, 1, 2, 3, 4, 5, 6, 7, 8, 9},
        ytick={0.76,0.77,...,0.83}
                ]
        \addplot[smooth,mark=*,blue] plot coordinates {
            (0,0.7700286279133701)
            (1,0.8071784439315165)
            (2,0.819403997499715)
            (3,0.8239669705573398)
            (4,0.8240654953735065)
            (5,0.823491266402737)
            (6,0.8209177960202888)
            (7,0.8187754799891874)
            (8,0.82)
            (9,0.819403997499715)
        };
        \end{axis}
	\end{tikzpicture}
	\caption{Text encoder training results}
	\label{fig_encoder_training_result}
\end{figure}

As shown in figure \ref{fig_encoder_training_result}, our text encoder reaches the best score at epoch 3 within 0.82 on cosine pearson. This means our text encoder can produce text embedding with higher cosine similarity on the similar sentence and vice versa.

\begin{center}
    \begin{table}[t]
    \caption{Evaluation results both on Inception Score and Fréchet inception distance compared with other English text-to-image generation models on CUB datasets}
    \begin{center}
    \begin{tabular}{|l|c|c|}
    \hline
    \textbf{Method} & \textbf{Inception Score} & \textbf{FID} \\
    \hline
    AttGAN \cite{Xu2018} & $4.36 \pm 0.03$ & 23.98 \\
    StackGAN \cite{Zhang2019} & $4.04 \pm 0.05$ & \textbf{15.30} \\
    MirrorGAN \cite{Qiao2019} & $4.56 \pm 0.05$ & - \\
    CycleGAN+BERT \cite{Tsue2020} & \textbf{5.92} & - \\
    SSA-GAN \cite{Hu2021} & $5.17 \pm 0.08$ & 15.61 \\
    \hline
    Our Unconditional & $4.75 \pm 0.19$ & 99.12 \\
    Our Conditional & $4.76 \pm 0.43$ & $46.401$ \\
    \hline
    \end{tabular}
    
    \label{quantitative_evaluation_results}
    \end{center}
    \end{table}
\end{center}


The fréchet inception distance and inception score of our model compared with other English text-to-image synthesis are shown in Table \ref{quantitative_evaluation_results}. Our model easily beats StackGAN, AttentionGAN, and MirrorGAN on the inception score, which means our model can generate high-quality objects. However, our FID is higher than other models, which means our models, compared to the training data, have different data distribution and makes our generated images different from our datasets.

\bigskip

The Inception Score for our unconditional GAN is slightly lower than our conditional GAN but has a much higher FID. This means our unconditional and conditional GAN has the same objectness but different image data distribution against the datasets.

\begin{center}
    \begin{table}[t]
    \caption{Examples of images generated by our Unconditional and Conditional GAN. The input sentences in English is "A red bird perched on a tree branch", "This bird has yellow wing and black beak", "A bird with blue wing and red tail", "A blackbird perched on a tree", and "A yellow bird on the water"}
    \begin{center}
    \begin{tabular}{lcc}
    \hline
    \textbf{Sentence} & \textbf{Unconditional} & \textbf{Conditional} \\
    \hline
   \makecell[bl]{\scriptsize seekor burung \\\scriptsize merah yang \\\scriptsize hinggap di \\\scriptsize cabang pohon} & \includegraphics[width=0.4\textwidth]{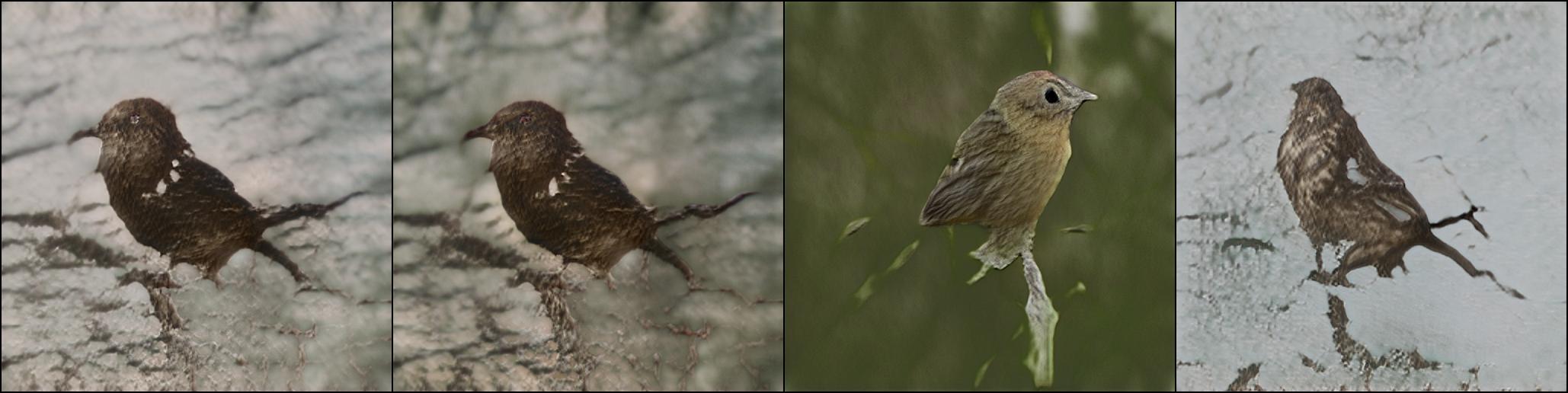} & \includegraphics[width=0.4\textwidth]{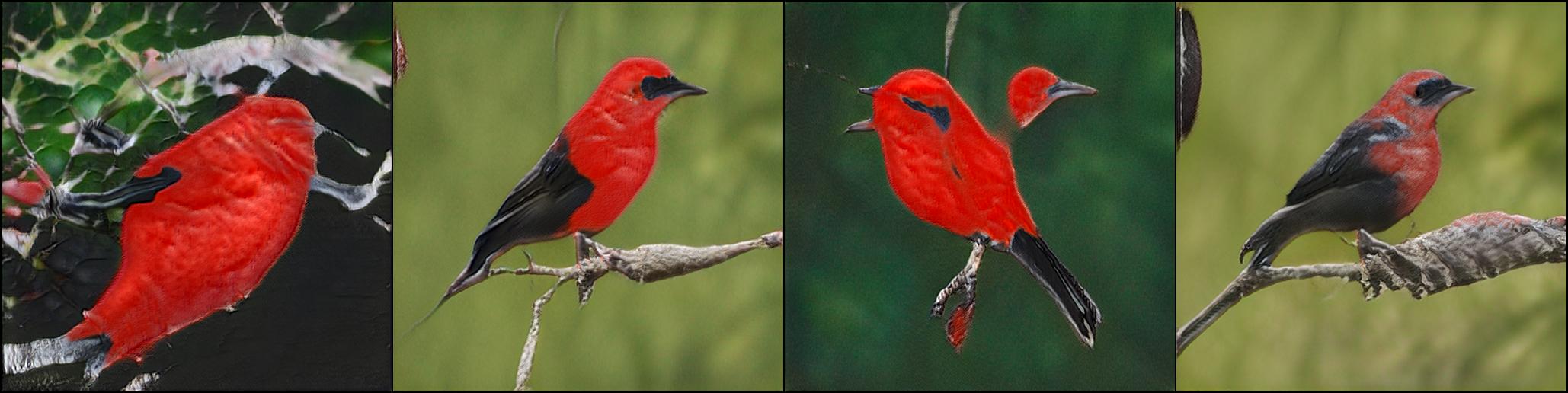} \\
 \hline
 \makecell[bl]{\scriptsize burung ini \\\scriptsize memiliki sayap \\\scriptsize berwarna kuning \\\scriptsize dengan paruh \\\scriptsize berwara hitam} & \includegraphics[width=0.4\textwidth]{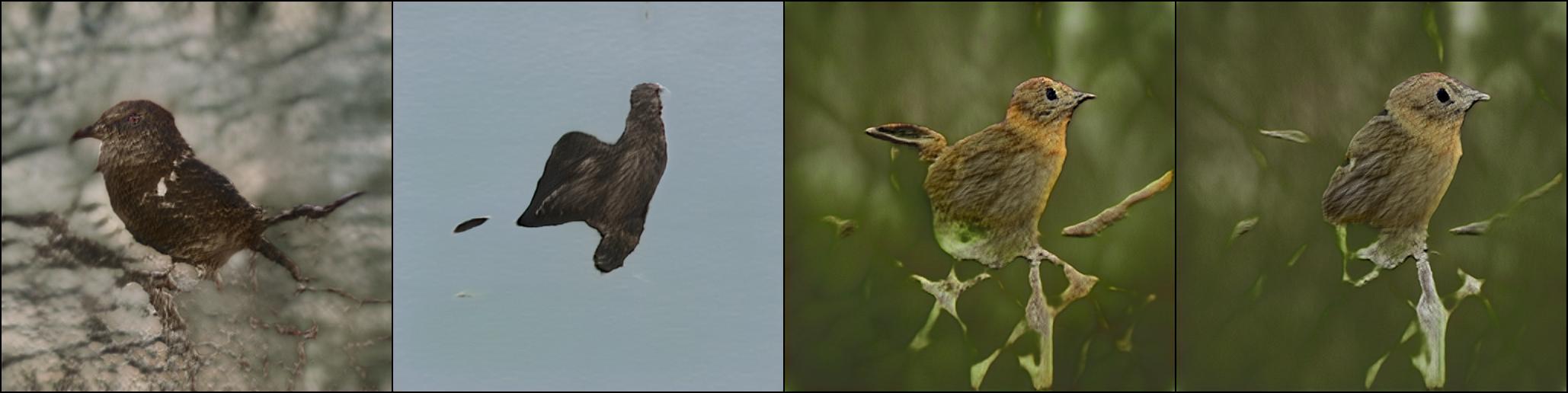} & \includegraphics[width=0.4\textwidth]{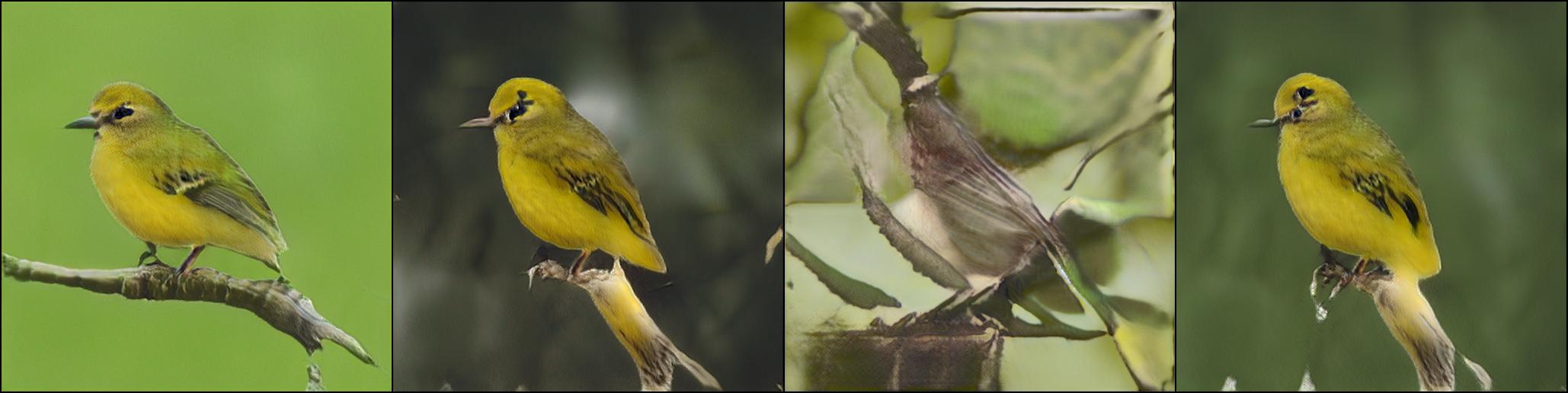} \\
 \hline
 \makecell[bl]{\scriptsize seekor burung \\\scriptsize dengan sayap \\\scriptsize berwarna biru \\\scriptsize dan ekor \\\scriptsize berwarna merah} & \includegraphics[width=0.4\textwidth]{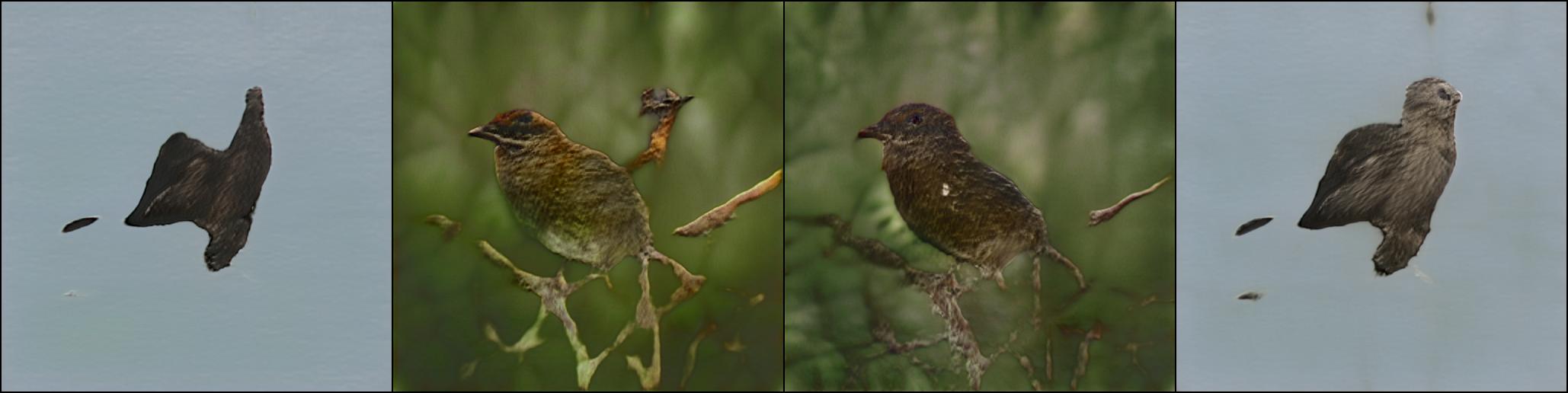} & \includegraphics[width=0.4\textwidth]{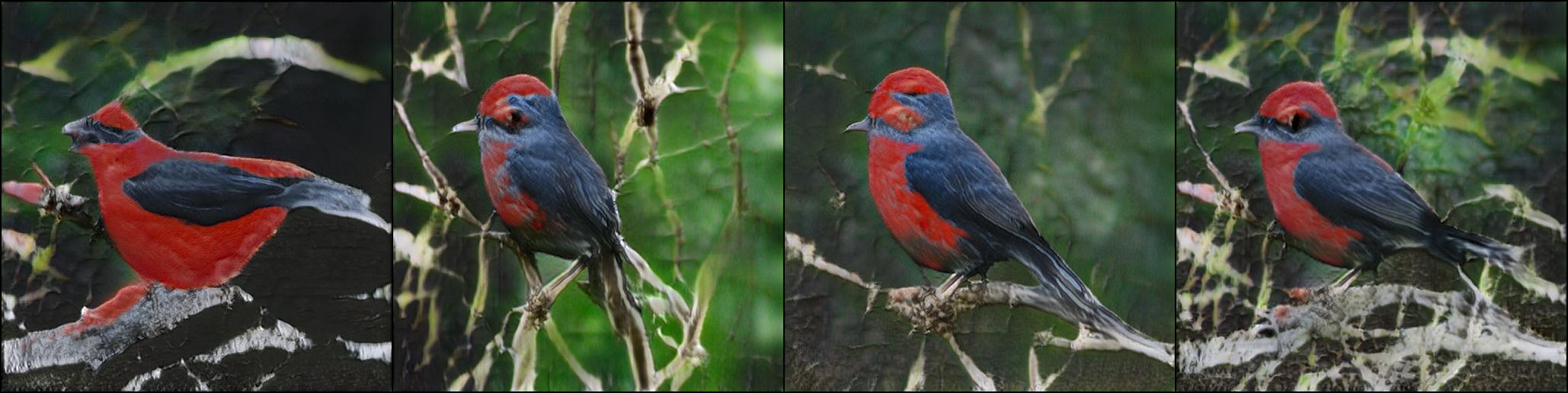} \\
 \hline
 \makecell[bl]{\scriptsize seekor burung \\\scriptsize hitam sedang \\\scriptsize bertengger di \\\scriptsize atas pohon} & \includegraphics[width=0.4\textwidth]{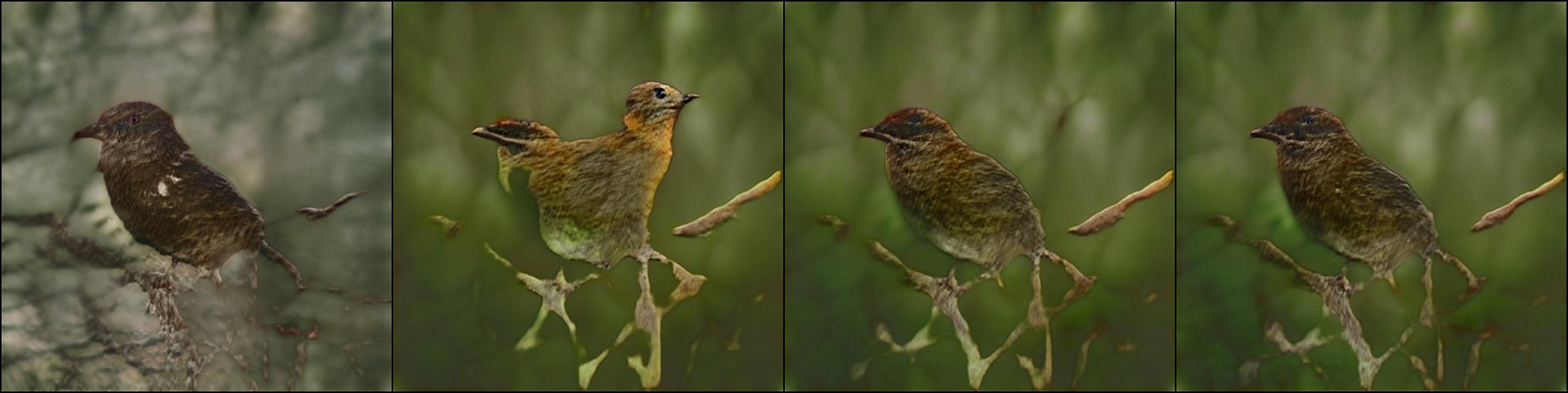} & \includegraphics[width=0.4\textwidth]{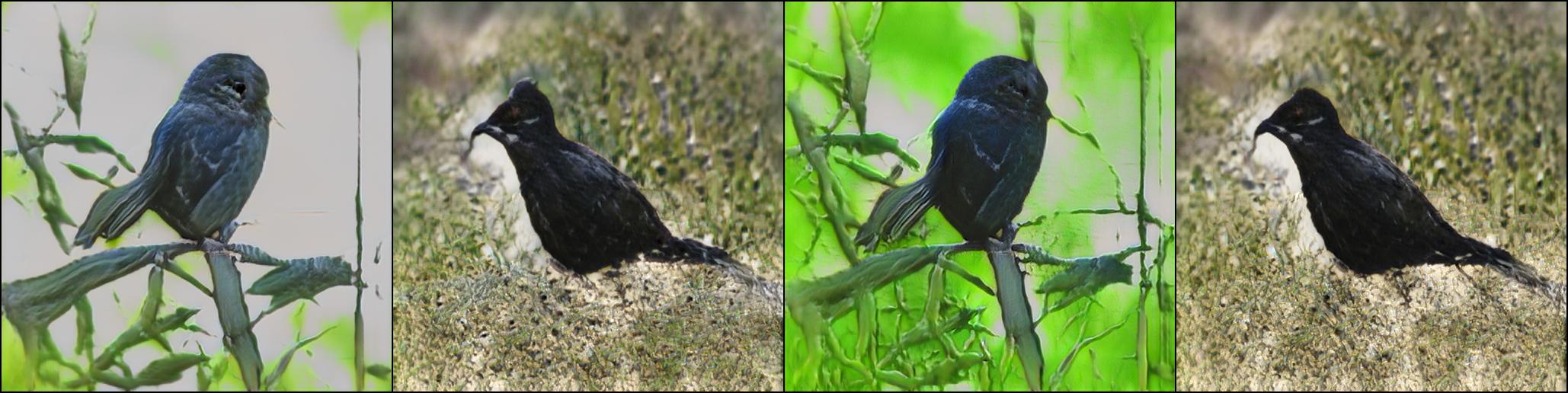} \\
 \hline
 \makecell[bl]{\scriptsize seekor burung \\\scriptsize kuning di \\\scriptsize atas air} & \includegraphics[width=0.4\textwidth]{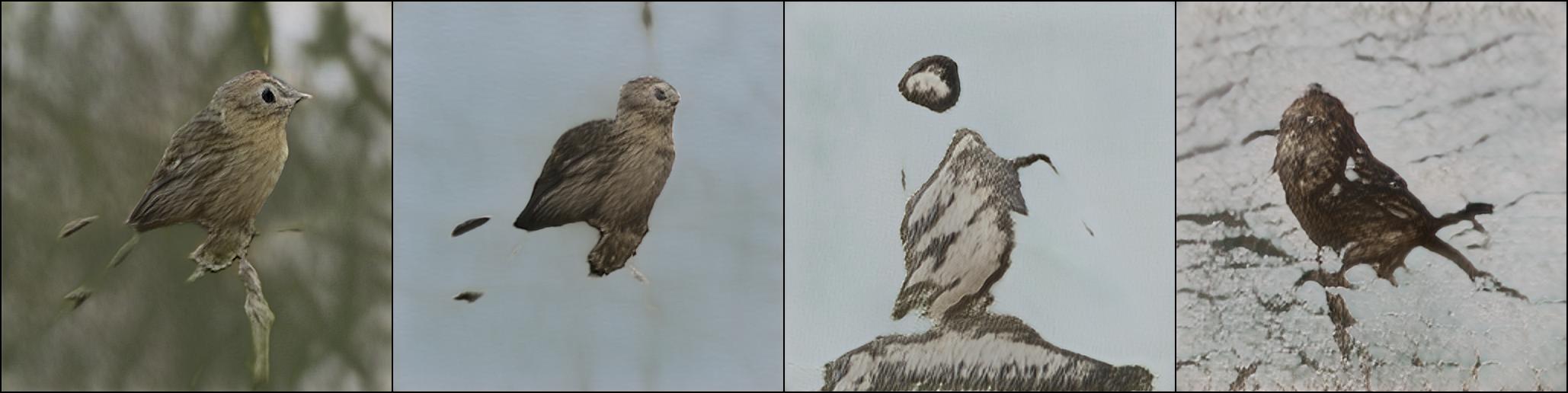} & \includegraphics[width=0.4\textwidth]{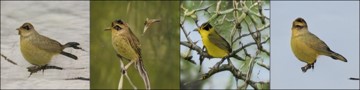} \\
    \hline
    \end{tabular}
    \label{example_generated_images}
    \end{center}
    \end{table}
\end{center}

As shown in Table \ref{example_generated_images}, we find out that our unconditional GAN is on collapse modes. This is a common problem on generative adversarial networks because the generator always tries to produce a similar image to fool the discriminator. This can happen when the discriminator cannot distinguish between fake and real images from different inputs. On the other hand, the conditional GAN can fix this problem using the conditioning augmentation from CA-net as input for the discriminator.

\bigskip

The conditional GAN also produces some novel images. The generator tries to generate a red bird with blue wings on the third output, which does not exist in the datasets. However, there is a red bird with black wings, so the generator is trying to change the wing color to blue. That makes the wing dark blue. Furthermore, there is no yellow bird on the water in the dataset, but the generator is trying to output that in the last sentence. So the generator is trying to change the existing yellow bird background with something like water.

\begin{center}
    \begin{table}[htbp]
    \caption{The mean opinion score for both our unconditional and conditional GAN}
    \begin{center}
    \begin{tabular}{|l|l|}
    \hline
    \textbf{Method} & \textbf{Mean Opinion Score} \\
    \hline
    Our Unconditional & 1.41 \\
    Our Conditional & \textbf{3.22} \\
    \hline
    \end{tabular}
    
    \label{qualitative_evaluation_results}
    \end{center}
    \end{table}
\end{center}


In order to perform a qualitative evaluation, we conduct a survey and average the results. As shown in Table \ref{qualitative_evaluation_results}, our conditional GAN easily beats our unconditional GAN. The mean opinion score of the conditional GAN is 3.22, which means our models are still acceptable by humans.

\section{Conclusion}
This research investigates the text-to-image synthesis performance in different languages, especially Bahasa Indonesia. To break through the gap between natural language and vision, we use the current state-of-the-art sentence embedding, Sentence BERT as the text encoder. In order to generate photo-realistic images within minimal training effort, we use FastGAN as the image generator. We implement the Conditioning Augmentation network to make the generated images more diverse and use its output as input for generator and discriminator. This makes our conditional GAN perform superior to generating novel images. Our proposed architecture can generate high-resolution images using Bahasa Indonesia as the input language.

\bigskip

For future works, we suggest properly translating the datasets to produce a high-quality language model. We find out that text embeddings play an essential role in generating the image details. Implementing weighted sum on the discriminator loss also can produce high-quality images. AttentionGAN \cite{Xu2018} become another alternative to produce high-quality details. Using another challenging dataset such as COCO datasets \cite{lin2014microsoft} or even ImageNet datasets can evaluate the model performance on different image distributions. It would be helpful to train the model longer to increase the inception score and FID. Trying other GAN variants also can produce different image resolutions, such as StyleGAN2 \cite{Karras2020}.





\bibliographystyle{dcu}
\bibliography{IJAMAS_example}

@incollection{goodfellow14,
title = {Generative Adversarial Nets},
author = {Goodfellow, Ian and Pouget-Abadie, Jean and Mirza, Mehdi and Xu, Bing and Warde-Farley, David and Ozair, Sherjil and Courville, Aaron and Bengio, Yoshua},
booktitle = {Advances in Neural Information Processing Systems 27},
editor = {Z. Ghahramani and M. Welling and C. Cortes and N. D. Lawrence and K. Q. Weinberger},
pages = {2672--2680},
year = {2014},
publisher = {Curran Associates, Inc.},
url = {http://papers.nips.cc/paper/5423-generative-adversarial-nets.pdf}
}

@article{Reed2016,
archivePrefix = {arXiv},
arxivId = {1605.05396},
author = {Reed, Scott and Akata, Zeynep and Yan, Xinchen and Logeswaran, Lajanugen and Schiele, Bernt and Lee, Honglak},
eprint = {1605.05396},
isbn = {9781510829008},
journal = {33rd International Conference on Machine Learning, ICML 2016},
pages = {1681--1690},
title = {{Generative adversarial text to image synthesis}},
volume = {3},
year = {2016}
}

@article{Mansimov2016,
archivePrefix = {arXiv},
arxivId = {1511.02793},
author = {Mansimov, Elman and Parisotto, Emilio and Ba, Jimmy Lei and Salakhutdinov, Ruslan},
eprint = {1511.02793},
journal = {4th International Conference on Learning Representations, ICLR 2016 - Conference Track Proceedings},
pages = {1--12},
title = {{Generating images from captions with attention}},
year = {2016}
}

@article{Tsue2020,
archivePrefix = {arXiv},
arxivId = {2003.12137},
author = {Tsue, Trevor and Li, Jason and Sen, Samir},
eprint = {2003.12137},
issn = {23318422},
journal = {arXiv},
title = {{Cycle Text-to-Image GAN with BERT}},
year = {2020}
}

@article{Zhang2019,
archivePrefix = {arXiv},
arxivId = {1710.10916},
author = {Zhang, Han and Xu, Tao and Li, Hongsheng and Zhang, Shaoting and Wang, Xiaogang and Huang, Xiaolei and Metaxas, Dimitris N.},
doi = {10.1109/TPAMI.2018.2856256},
eprint = {1710.10916},
issn = {19393539},
journal = {IEEE Transactions on Pattern Analysis and Machine Intelligence},
number = {8},
pages = {1947--1962},
pmid = {30010548},
title = {{StackGAN++: Realistic Image Synthesis with Stacked Generative Adversarial Networks}},
volume = {41},
year = {2019}
}

@techreport{WahCUB_200_2011,
	Title = {{The Caltech-UCSD Birds-200-2011 Dataset}},
	Author = {Wah, C. and Branson, S. and Welinder, P. and Perona, P. and Belongie, S.},
	Year = {2011},
	Institution = {California Institute of Technology},
	Number = {CNS-TR-2011-001}
}

@article{Qiao22019,
author = {Qiao, Tingting and Zhang, Jing and Xu, Duanqing and Tao, Dacheng},
issn = {10495258},
journal = {Advances in Neural Information Processing Systems},
number = {NeurIPS},
pages = {1--11},
title = {{Learn, imagine and create: Text-to-image generation from prior knowledge}},
volume = {32},
year = {2019}
}

@article{Xu2018,
archivePrefix = {arXiv},
arxivId = {1711.10485},
author = {Xu, Tao and Zhang, Pengchuan and Huang, Qiuyuan and Zhang, Han and Gan, Zhe and Huang, Xiaolei and He, Xiaodong},
doi = {10.1109/CVPR.2018.00143},
eprint = {1711.10485},
isbn = {9781538664209},
issn = {10636919},
journal = {Proceedings of the IEEE Computer Society Conference on Computer Vision and Pattern Recognition},
pages = {1316--1324},
title = {{AttnGAN: Fine-Grained Text to Image Generation with Attentional Generative Adversarial Networks}},
year = {2018}
}

@article{Devlin2019,
archivePrefix = {arXiv},
arxivId = {1810.04805},
author = {Devlin, Jacob and Chang, Ming Wei and Lee, Kenton and Toutanova, Kristina},
eprint = {1810.04805},
isbn = {9781950737130},
journal = {NAACL HLT 2019 - 2019 Conference of the North American Chapter of the Association for Computational Linguistics: Human Language Technologies - Proceedings of the Conference},
number = {Mlm},
pages = {4171--4186},
title = {{BERT: Pre-training of deep bidirectional transformers for language understanding}},
volume = {1},
year = {2019}
}

@article{Qiao2019,
archivePrefix = {arXiv},
arxivId = {1903.05854},
author = {Qiao, Tingting and Zhang, Jing and Xu, Duanqing and Tao, Dacheng},
doi = {10.1109/CVPR.2019.00160},
eprint = {1903.05854},
isbn = {9781728132938},
issn = {10636919},
journal = {Proceedings of the IEEE Computer Society Conference on Computer Vision and Pattern Recognition},
pages = {1505--1514},
title = {{Mirrorgan: Learning text-to-image generation by redescription}},
volume = {2019-June},
year = {2019}
}

@article{Liu2021,
archivePrefix = {arXiv},
arxivId = {2103.14030},
author = {Liu, Ze and Lin, Yutong and Cao, Yue and Hu, Han and Wei, Yixuan and Zhang, Zheng and Lin, Stephen and Guo, Baining},
eprint = {2103.14030},
title = {{Swin Transformer: Hierarchical Vision Transformer using Shifted Windows}},
url = {http://arxiv.org/abs/2103.14030},
year = {2021}
}

@article{Hu2021,
archivePrefix = {arXiv},
arxivId = {2104.00567},
author = {Hu, Kai and Liao, Wentong and Yang, Michael Ying and Rosenhahn, Bodo},
eprint = {2104.00567},
number = {1},
title = {{Text to Image Generation with Semantic-Spatial Aware GAN}},
url = {http://arxiv.org/abs/2104.00567},
year = {2021}
}

@article{Liuf2021,
archivePrefix = {arXiv},
arxivId = {2101.04775},
author = {Liu, Bingchen and Zhu, Yizhe and Song, Kunpeng and Elgammal, Ahmed},
eprint = {2101.04775},
pages = {1--22},
title = {{Towards Faster and Stabilized GAN Training for High-fidelity Few-shot Image Synthesis}},
url = {http://arxiv.org/abs/2101.04775},
year = {2021}
}

@article{Reimers2020,
archivePrefix = {arXiv},
arxivId = {1908.10084},
author = {Reimers, Nils and Gurevych, Iryna},
doi = {10.18653/v1/d19-1410},
eprint = {1908.10084},
isbn = {9781950737901},
journal = {EMNLP-IJCNLP 2019 - 2019 Conference on Empirical Methods in Natural Language Processing and 9th International Joint Conference on Natural Language Processing, Proceedings of the Conference},
pages = {3982--3992},
title = {{Sentence-BERT: Sentence embeddings using siamese BERT-networks}},
year = {2020}
}

@article{Zhang2018,
archivePrefix = {arXiv},
arxivId = {1801.03924},
author = {Zhang, Richard and Isola, Phillip and Efros, Alexei A. and Shechtman, Eli and Wang, Oliver},
doi = {10.1109/CVPR.2018.00068},
eprint = {1801.03924},
isbn = {9781538664209},
issn = {10636919},
journal = {Proceedings of the IEEE Computer Society Conference on Computer Vision and Pattern Recognition},
number = {1},
pages = {586--595},
title = {{The Unreasonable Effectiveness of Deep Features as a Perceptual Metric}},
year = {2018}
}

@article{Heusel2017,
archivePrefix = {arXiv},
arxivId = {1706.08500},
author = {Heusel, Martin and Ramsauer, Hubert and Unterthiner, Thomas and Nessler, Bernhard and Hochreiter, Sepp},
eprint = {1706.08500},
issn = {10495258},
journal = {Advances in Neural Information Processing Systems},
number = {Nips},
pages = {6627--6638},
title = {{GANs trained by a two time-scale update rule converge to a local Nash equilibrium}},
volume = {2017-December},
year = {2017}
}

@article{Zhu2017,
archivePrefix = {arXiv},
arxivId = {1703.10593},
author = {Zhu, Jun Yan and Park, Taesung and Isola, Phillip and Efros, Alexei A.},
doi = {10.1109/ICCV.2017.244},
eprint = {1703.10593},
isbn = {9781538610329},
issn = {15505499},
journal = {Proceedings of the IEEE International Conference on Computer Vision},
pages = {2242--2251},
title = {{Unpaired Image-to-Image Translation Using Cycle-Consistent Adversarial Networks}},
volume = {2017-Octob},
year = {2017}
}

@article{Karras2020,
archivePrefix = {arXiv},
arxivId = {1912.04958},
author = {Karras, Tero and Laine, Samuli and Aittala, Miika and Hellsten, Janne and Lehtinen, Jaakko and Aila, Timo},
doi = {10.1109/CVPR42600.2020.00813},
eprint = {1912.04958},
issn = {10636919},
journal = {Proceedings of the IEEE Computer Society Conference on Computer Vision and Pattern Recognition},
pages = {8107--8116},
title = {{Analyzing and improving the image quality of stylegan}},
year = {2020}
}

@misc{cahya,
  author = {Cahya Wirawan},
  title = {{Indonesian BERT base model (uncased)}},
  howpublished = "\url{https://huggingface.co/cahya/bert-base-indonesian-522M}",
  year = {2020}, 
  note = "[Online; accessed 30-August-2021]"
}

@article{He2016,
archivePrefix = {arXiv},
arxivId = {1512.03385},
author = {He, Kaiming and Zhang, Xiangyu and Ren, Shaoqing and Sun, Jian},
doi = {10.1109/CVPR.2016.90},
eprint = {1512.03385},
isbn = {9781467388504},
issn = {10636919},
journal = {Proceedings of the IEEE Computer Society Conference on Computer Vision and Pattern Recognition},
pages = {770--778},
title = {{Deep residual learning for image recognition}},
volume = {2016-Decem},
year = {2016}
}

@misc{lim2017geometric,
      title={Geometric GAN}, 
      author={Jae Hyun Lim and Jong Chul Ye},
      year={2017},
      eprint={1705.02894},
      archivePrefix={arXiv},
      primaryClass={stat.ML}
}

@misc{lin2014microsoft,
  author = {Lin, Tsung-Yi and Maire, Michael and Belongie, Serge and Bourdev, Lubomir and Girshick, Ross and Hays, James and Perona, Pietro and Ramanan, Deva and Zitnick, C. Lawrence and Dollár, Piotr},
  description = {Microsoft COCO: Common Objects in Context},
  note = {cite arxiv:1405.0312Comment: 1) updated annotation pipeline description and figures; 2) added new  section describing datasets splits; 3) updated author list},
  title = {Microsoft COCO: Common Objects in Context},
  url = {http://arxiv.org/abs/1405.0312},
  year = 2014
}


\end{document}